\documentclass[conference]{IEEEtran}
\IEEEoverridecommandlockouts
\usepackage{cite}
\usepackage{amsmath,amssymb,amsfonts}
\usepackage{algorithmic}
\usepackage{graphicx}
\usepackage{textcomp}
\usepackage{xcolor}
\usepackage[hyphens]{url}
\usepackage{hyperref} 
\def\BibTeX{{\rm B\kern-.05em{\sc i\kern-.025em b}\kern-.08em
    T\kern-.1667em\lower.7ex\hbox{E}\kern-.125emX}}

\usepackage{fancyvrb}
\usepackage{mdframed}
\usepackage{listings}
\usepackage[numbers]{natbib}
\usepackage{array}

\begin{document}

\title{Garbage in, garbage out: Zero-shot detection of crime using Large Language Models}

\author{
\IEEEauthorblockN{Anj Simmons, Rajesh Vasa}
\IEEEauthorblockA{Applied Artificial Intelligence Institute, 
Deakin University, Geelong, Australia\\
Email: \{a.simmons, rajesh.vasa\}@deakin.edu.au}
}

\maketitle

\begin{abstract}
This paper proposes exploiting the common sense knowledge learned by large language models to perform zero-shot reasoning about crimes given textual descriptions of surveillance videos. We show that when video is (manually) converted to high quality textual descriptions, large language models are capable of detecting and classifying crimes with state-of-the-art performance using only zero-shot reasoning. However, existing automated video-to-text approaches are unable to generate video descriptions of sufficient quality to support reasoning (garbage video descriptions into the large language model, garbage out).
\end{abstract}


\section{Introduction}

Intelligence and law enforcement agencies are tasked with detecting threats and preventing crime. Such agencies have access to increasing volumes of data; however, as the amount of information available greatly exceeds the capacity of humans available to monitor it, it is impossible to fully monitor this deluge of information and respond in a timely manner. Therefore, there is a need for more sophisticated techniques to ingest information and surface up just the cases that indicate a potential crime of a category relevant to the agency.


Automatically detecting crime poses a challenge, as while there are known categories of crime (abuse, arson, assault, burglary, stealing, vandalism, etc.) the ways in which each category of crime can be committed are diverse. Furthermore, the losses caused by crime follow a power law distribution \cite{Ilin2021} in which rare events (e.g., terrorist incidents) cause disproportionate losses. Contrast this to the assumption of supervised learning approaches, which perform well on categories with sufficient training data, but perform poorly on categories with limited training data. Although supervised learning approaches may still have a role to play in extracting specific features relevant to crime, e.g., the presence of a weapon, it is necessary to consider the broader context of the scenario to determine if a crime is occurring, which has traditionally required a human.

\begin{figure}
    \includegraphics[width=0.32\linewidth]{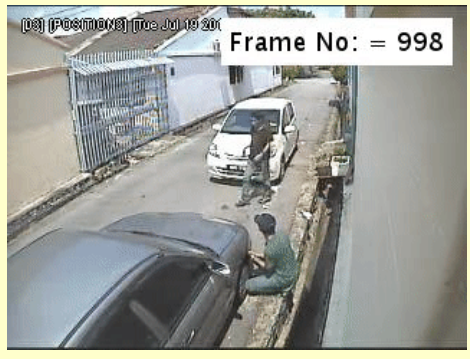}
    \includegraphics[width=0.32\linewidth]{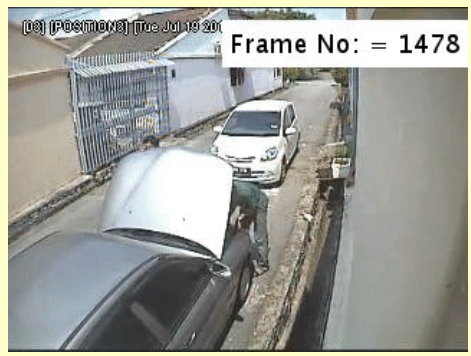}
    \includegraphics[width=0.32\linewidth]{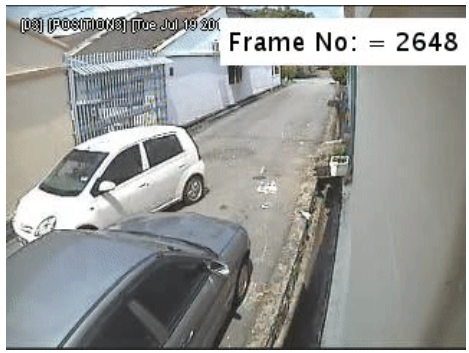}
    \small
    \textbf{1) Prompt:} \\
SURVEILLANCE VIDEO DESCRIPTION: \\
A white car approaches a silver car. Two people come out. One of them goes near the front wheel of the car. The other then opens the bonnet. After 30 seconds, the two people return to the white car. \\
\\
TASK: \\
List possible explanations, and categorise them as one of [Abuse, Arrest, Arson, Assault, Burglary, Explosion, Fighting, Road Accident, Robbery, Shooting, Shoplifting, Stealing, Vandalism, Normal]. Then determine which category is most likely. Let's think step by step. Finally, output one line containing a single category in quotes. Do not include anything other than the category on the final line.
\\
\\
    \textbf{2) GPT-4 Chain-of-Thought:} \\
Possible explanations: \\
1. The two people are helping the driver of the silver car with a mechanical issue [Normal] \\
2. The two people are attempting to steal parts from the silver car [Stealing] \\
3. The two people are tampering with the silver car to cause harm [Vandalism] \\
4. The two people are involved in a car theft [Robbery] \\
\\
Most likely category: ``Normal''
    \caption{Example application of proposed method for zero-shot detection of crime: 1) Video is converted to a prompt consisting of a textual description of the video along with a fixed task description. The surveillance video description in this example was written by a human, but we also explore approaches to automate this step. 2) GPT-4 is used to reason (i.e., produce a Chain-of-Thought) about possible explanations and determine the most likely category.
    }
    \label{fig:example}
\end{figure}

In this paper, we propose exploiting the zero-shot reasoning capabilities \cite{kojima2022large} of Large Language Models (LLMs) for the task of detecting and reasoning about crime. Zero-shot reasoning allows prompting the LLM to reason about whether a description of events suggests evidence of a crime, without the need to provide training examples (other than the corpus the LLM has been pre-trained on). LLMs capture common sense knowledge (albeit at a surface level) \cite{Zhou2020} which is important to support this reasoning process and avoid false positives, for example, it is not a crime to take an item from a shop if one pays for it before leaving. An example of the proposed approach is shown in \autoref{fig:example}.


To test the approach, we evaluate the ability of a state-of-the-art LLM (GPT-4 \cite{OpenAI2023}) to detect crimes given textual descriptions of events in real-world surveillance videos. For the purposes of this paper, we manually created descriptions of 168 surveillance videos. We also explored approaches for automatically generating descriptions from video, which would allow for a fully automated approach to detecting crime in surveillance video, but found that the quality of the automatically generated descriptions was insufficient for the LLM to accurately detect and reason about crimes (garbage in, garbage out).

The key contributions of this paper are:
\begin{enumerate}
    \item A dataset of textual descriptions derived from real-world surveillance videos which can be used to benchmark the ability of LLMs to reason about crime, and
    \item Identification of obstacles in existing video-to-text approaches that prevent a fully-automated approach.
\end{enumerate}

Data and code for this paper are publicly available online\footnote{\url{https://github.com/anjsimmo/zero-shot-crime-detection}}.

\section{Background and Related work}

Previous work has considered the task of activity recognition in videos. However, this requires large training datasets, which are not available in the case of rare types of crime. Previous work has also considered training multimodal models to support zero-shot reasoning about images and other input modalities. However, such models have limited transparency about which information in the images are used to inform the decision, which is essential to know and control in the case of crime detection.

\subsection{Activity Recognition}

Large-scale datasets have been collected for the task of describing activities in video, such as \textsc{VaTeX} \cite{Wang_2019_ICCV} (consisting of 41,250 videos with annotations). For the task of crime detection, the UCF-Crime dataset \cite{sultani2018real} consists of 128 hours of surveillance video obtained from YouTube and LiveLeak. However, existing activity recognition methods perform poorly on the UCF-Crime dataset, only achieving 28.4\% classification accuracy \cite{sultani2018real}. Furthermore, rare crime categories such as terrorism are not included. In contrast to previous crime detection approaches, this paper explores a zero-shot approach to circumvent the need for large training datasets.







\subsection{Multimodal Models}

Multimodal models are trained on two or more input modalities, such as both text and images. Such models can accept image inputs directly, allowing reasoning about content in images without the need for an intermediate step to first convert video to text. Prior work has also explored integrating LLMs with pre-trained image encoders to allow reasoning about the content of images \cite{liu2023llava}.

There are three reasons why we focus on reasoning about textual descriptions of videos in this paper rather than training a multimodal model to operate on the video directly. Firstly, using textual descriptions allows the use of existing state of the art LLMs, such as GPT-4 and derivatives of LLaMA, without the need to retrain or fine-tune them to accept new input modalities. Although the GPT-4 model itself is multimodal \cite{OpenAI2023}, OpenAI do not yet provide a way for the public to input images to GPT-4 via the API. Secondly, we may wish to incorporate information other than images in future, for example, descriptions of sounds detected or new kinds of sensors. When all information sources are represented in textual form, it is trivial to integrate new information. Thirdly, textual descriptions provide a way to restrict which information the LLM has access to by censoring details the LLM should not use in its decision, such as race and gender. Inspecting the chain-of-thought produced by the LLM for bias is insufficient, as LLMs may produce unfaithful explanations that do not reveal the underlying factors that influenced the decision \cite{Turpin2023}. Censoring these details is more difficult when the input includes images/videos.

\begin{figure}
    \centering
    \includegraphics[width=0.85\linewidth]{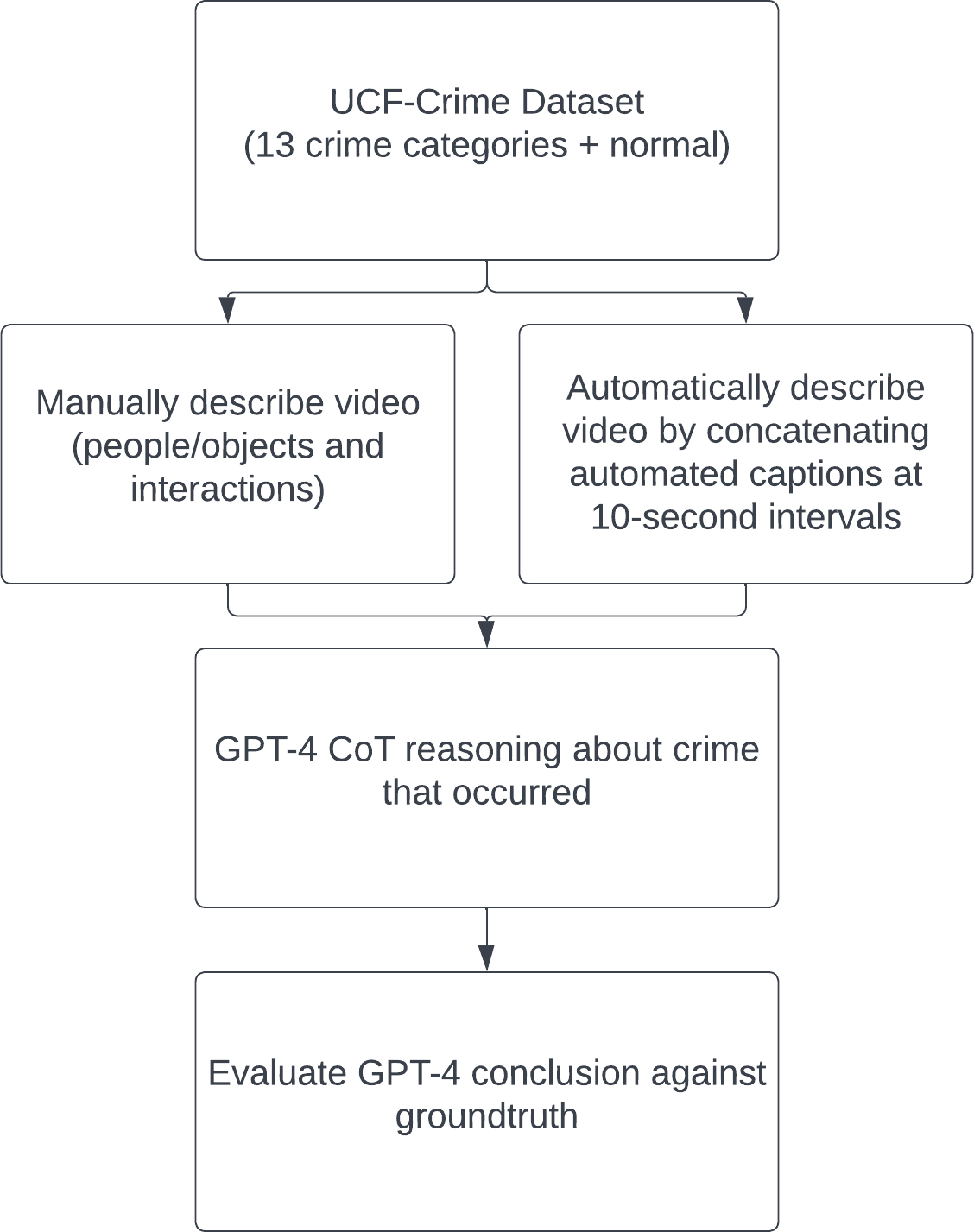}
    \caption{We provide descriptions of people/objects and interactions within the UCF-Crime dataset, then evaluate the ability of GPT-4 chain-of-thought reasoning to infer the category of crime.}
    \label{fig:process}
\end{figure}

\section{Method}

This section explains the method by which we converted surveillance videos into textual descriptions and evaluated the ability of GPT-4 to detect and reason about crime. An overview of the process is shown in \autoref{fig:process}.

\subsection{Dataset}

We test our approach on the UCF-Crime dataset of surveillance videos \cite{sultani2018real}. The authors of the UCF-Crime \cite{sultani2018real} propose two tasks. The first task is detection of anomalous events in surveillance videos, on which the original paper scores 75.41 AUC, and current state of the art scores 86.98 AUC\footnote{\url{https://paperswithcode.com/sota/anomaly-detection-in-surveillance-videos-on}}. However, this first task only involves identifying which frames relate to an anomalous event, not reasoning about the type of anomalous event. The second task is anomalous activity recognition. This involves determining the category of crime (if any) that occurs in a surveillance video. The authors of the UCF-Crime dataset explain that state of the art activity recognition methods perform poorly, only achieving 28.4\% accuracy. This second task is the focus of this paper.

\subsection{Human Captions}

The first author manually wrote descriptions of people, objects, and their interactions in a sample of 12 surveillance videos from 14 categories (13 crime categories + normal), resulting in a dataset of 168 surveillance video descriptions. While the author was aware of the ground truth crime category for videos in the UCF-Crime dataset, the descriptions were written to be as objective as possible rather than suggesting a particular interpretation of the events taking place. The manually written descriptions do not include race or gender.

An example of a manually created description is: ``A white car approaches a silver car. Two people come out. One of them goes near the front wheel of the car. The other then opens the bonnet. After 30 seconds, the two people return to the white car.''

\subsection{GPT-4 Prompt}

We prompt GPT-4, a state of the art LLM, to reason about the most likely category of crime given the textual description of the surveillance video. We include the phrase ``let's think step by step'' in the prompt to trigger zero-shot reasoning \cite{kojima2022large}. The full prompt is:

\begin{mdframed}
SURVEILLANCE VIDEO DESCRIPTION: \\
A white car approaches a silver car ... \\
\\
TASK: \\
List possible explanations, and categorise them as one of [Abuse, Arrest, Arson, Assault, Burglary, Explosion, Fighting, Road Accident, Robbery, Shooting, Shoplifting, Stealing, Vandalism, Normal]. Then determine which category is most likely. Let's think step by step. Finally, output one line containing a single category in quotes. Do not include anything other than the category on the final line.
\end{mdframed}


GPT-4 reasons about possible explanations for the observed events, followed by what it determines to be the most likely category. An example response from GPT-4 is:

\begin{mdframed}
Possible explanations: \\
1. The two people are helping the driver of the silver car with a mechanical issue [Normal] \\
2. The two people are attempting to steal parts from the silver car [Stealing] \\
3. The two people are tampering with the silver car to cause harm [Vandalism] \\
4. The two people are involved in a car theft [Robbery] \\
\\
Most likely category: ``Normal''
\end{mdframed}

We evaluate how accurately GPT-4 is able to recover the ground truth label for the video.

\subsection{Automated Captions}

We also explore the accuracy with which GPT-4 can recover the ground truth on automatically generated captions of videos. To automatically generate a description of the video, we sample a frame every 10 seconds, pass it through a image-to-text model, then concatenate the results along with time stamps. The choice to sample a frame every 10 seconds was to balance the need to capture key changes in the video while avoiding long and repetitive descriptions that may exceed the input token limit of GPT-4. The method for converting these descriptions to prompts and evaluating the classification accuracy is identical to the process for human captions.

\subsubsection{GIT Captions}

To generate image captions, we used a pre-trained Generative Image-to-text Transformer (GIT) \cite{Wang2022a}, specifically, \texttt{git-large-coco}\footnote{\url{https://huggingface.co/microsoft/git-large-coco}} which has been fine-tuned by Microsoft on the COCO dataset.

An example of the final description generated for a video by this process is: ``10 sec: a car is seen parked on the street. 20 sec: a car is seen parked on the street. 30 sec: a car is seen passing another car. 40 sec: a man is sitting on the ground next to a car. 50 sec: a car is parked on the street and another car is parked behind it. 60 sec: a man is opening the trunk of a car. 70 sec: a man is trying to get a car out of the back of a car. 80 sec: a car is parked in front of a house. 90 sec: a car is parked on the street. 100 sec: a car is parked on the street. 110 sec: the car is parked on the street. 120 sec: a car is seen driving down a street.''

\subsubsection{LLaVA Descriptions}

Large Language and Vision Assistant (LLaVA) \cite{liu2023llava} is a visual instruction tuned version of LLaMa. Specifically, we use the 13 billion parameter version, \texttt{13b-v0}, which was the largest and most recent release of the LLaVA model at the time of conducting the experiment. To generate descriptions, we prompt LLaVA with an image from the video and the question ``What is it?''

An example of the final description generated for a video by this process is: ``10 sec: In the image, there is a black car parked on a street next to a building. The car appears to be parked in a parking space, and there are potted plants nearby. The car is facing a house with a fence and a gate. There is also a person standing on the sidewalk, possibly observing the car or the surrounding area. To assist the user further, more context or specific questions about the scene would be needed. 20 sec: In the image, there is a gray car parked on the side of the street, and a white van is driving down the street. The scene takes place on a residential street with a fence and a house in the background. To assist you better, I would need more information about the situation or a specific question related to the image. 30 sec: ...''

\subsubsection{YOLO-v8 + ByteTrack}

We make use of a YOLO-based object tracking library\footnote{\url{https://github.com/mikel-brostrom/yolo_tracking}} to track people and objects in surveillance videos. Specifically, we use \texttt{yolov8x}\footnote{\url{https://github.com/ultralytics/ultralytics}} (the largest and most accurate version of the model) combined with the ByteTrack \cite{zhang2022bytetrack} tracking method.

The tracks are updated every frame (30 frames per second) to support keeping track of identities; however, we only output the current state every 10 seconds. To convert this to text, for each tracked object, we state the object class (person, car, etc.), identity of the tracked object (e.g. ``car 2'') and position (e.g. ``bottom-left'').

An example of the final description generated for a video by this process is: ``0 sec: car 1 is at the bottom-left of the image. 10 sec: car 1 is at the bottom-left of the image. car 2 is at the top-middle of the image. 20 sec: car 1 is at the bottom-left of the image. car 2 is at the middle of the image. 30 sec: car 1 is at the bottom-left of the image. car 2 is at the middle of the image. person 3 is at the bottom-middle of the image. person 5 is at the middle of the image. 40 sec: car 1 is at the bottom-left of the image. car 2 is at the middle of the image. person 5 is at the middle of the image. person 6 is at the bottom-middle of the image. 50 sec: ...''





\section{Results}

\begin{table}[]
\caption{\label{tab:results}Classification accuracy}
\centering
\begin{tabular}{lr}
\hline
\textbf{Method}                & \textbf{Accuracy} \\ \hline
Random Baseline                & 7.1\%                            \\
TCNN Baseline (Sultani et al.) & 28.4\%                           \\
\textbf{GPT-4 + Human Captions}         & \textbf{58.7}\%                           \\ 
GPT-4 + GIT Captions           & 11.4\%                            \\
GPT-4 + LLaVA Descriptions         & 10.2\%                            \\
GPT-4 + YOLO-v8 + ByteTrack    & 7.1\%                            \\ \hline
\end{tabular}
\end{table}

We tested each method on 12 videos from 14 categories (168 videos total) and report the classification accuracy in \autoref{tab:results}. For comparison, we also include a random baseline (1/14), and the Tube Convolutional Neural Network (TCNN) baseline reported by \citet{sultani2018real} in the UCF-Crime paper.

In cases where GPT-4 was unable to process the input (e.g. exceeded token length) or did not output a valid response (i.e. the final line of output was not a valid category in the expected format) we exclude these from the accuracy calculation. There was 1 case of invalid output for GPT-4 + Human captions, 1 case of invalid output for GPT-4 + GIT Captions, and 2 cases of input that could not be processed due to exceeding token length for GPT-4 + LLaVA Captions. There were no cases of invalid input or output for GPT-4 + YOLO-v8 + ByteTrack.



\section{Discussion}

Our results show that while GPT-4 was able to determine the crime category with state of the art performance when provided a human generated caption of the video, it performed poorly when provided with automatically generated captions. In the rest of this section, we elaboration on the limitations of automatic caption generation approaches that need to be overcome to support a fully automated approach to crime detection.

\subsection{Image captions lack detail}

\begin{figure}
    \centering
    \includegraphics[width=0.8\linewidth]{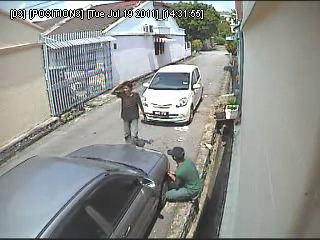}
    \caption{GIT Caption: ``a man is sitting on the ground next to a car.''}
    \label{fig:git-caption-failure}
\end{figure}

Reasoning about crime requires details of who did what. However, image captioning models only describe the scene at a high level. For example, consider the caption generated for \autoref{fig:git-caption-failure}. The image captioning model correctly identifies that the image contains a man sitting on the ground next to a car. However, it provides no details about which man and which car. Without this detail, there is insufficient information to reason about whether the man is sitting on the ground to repair their own car, or is sitting on the ground to steal something from someone else's car.

\subsection{LLM based vision models hallucinate}

\begin{figure}
    \centering
    \includegraphics[width=0.8\linewidth]{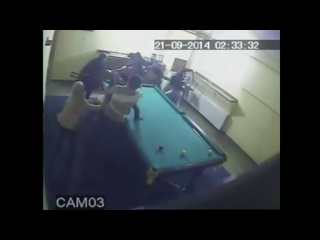}
    \caption{LLaVA Description: ``In the image, a group of young men is playing pool together in a room. There are at least seven people present, with some standing and others sitting around the pool table. The players are engaged in a game of pool, with the cue ball visible on the table.  The room appears to be a casual gathering space, as there are chairs and a bench placed around the area. Additionally, there is a microwave in the room, suggesting that it might be a part of a larger living space, such as a dormitory or a recreational area. The overall atmosphere seems to be one of enjoyment and camaraderie among the players.''}
    \label{fig:llava-failure}
\end{figure}

In contrast to image captioning models, LLM based vision models are capable of generating detailed descriptions of images, but may hallucinate about objects present and actions being performed. Furthermore, the descriptions they generate are biased towards a particular interpretation of the scenario, an example of which is shown in \autoref{fig:llava-failure}. Reasoning about crime requires objective descriptions.

\subsection{Object tracking algorithms cannot maintain identity of objects over long time periods}

Reasoning about crime requires linking the identity of actors across time. For example, it is not a crime for an actor to take an item from a store if they pay for it before leaving.

However, in our experiments, we observed that object tracking was unable to maintain a constant identity for people and objects over long time periods. For example, in a video where there were only two people, the description generated by applying object tracking refers to ``person 3'', ``person 5'', ``person 6'' and ``person 8'', making it difficult to link the actions of people across time.

\subsection{Curated object detection datasets do not include weapons}

Large benchmark datasets for object detection, such as COCO, do not include weapons. While it is understandable that technology companies that curate datasets for training machine learning algorithms may want to distance themselves from undesirable uses of AI, if we wish for AI systems to be able to help prevent violence, then it is important for datasets curators to include depictions of weapons and violence.

\section{Threats to validity}

GPT-4 may have already have seen images from UCF-Crime videos in its training data. However, we test on textual descriptions of the videos rather than the videos themselves, and these textual descriptions have not been released before. As such, it is unlikely that the performance of GPT-4 reported in this paper is a result of overfitting to training data.

The author was aware of the ground-truth category when creating the human captions for videos, which may have biased the descriptions. Furthermore, in busy scenes, it was not practical to describe every action taking place, hence the descriptions may be biased towards just describing the actions of relevance to the crime. As such, the performance of GPT-4 on human captions should be taken only as an indicator of what is possible, and may not be possible to fully automate even if the obstacles raised in this paper can be overcome.

\section{Conclusion}

This paper demonstrated that with high quality textual descriptions, large language models are capable of detecting and classifying crimes with state-of-the-art performance using only zero-shot reasoning. Unfortunately, existing automated video-to-text approaches were unable to generate video descriptions of sufficient quality to support reasoning, thus fully automated detection of crime is not yet possible. The failure of these approaches to generate descriptions suitable for reasoning about crime indicates that such models are not as general purpose as widely perceived, and that these models require domain adaptation for downstream tasks. Future research is needed to overcome the loss of objective detail that occurs during the video-to-text conversion process.

\section*{Acknowledgment}

This paper was supported by research funding from the National Intelligence Postdoctoral Grant program (NIPG-2021-006).

\bibliographystyle{IEEEtranN}
\bibliography{refs}


\end{document}